\documentclass[11pt,twocolumn]{article}
\usepackage{authblk}

\usepackage{xurl}  
\usepackage[hidelinks]{hyperref}
\usepackage{graphicx}
\usepackage{subcaption}
\usepackage{color}
\usepackage[table,dvipsnames]{xcolor}
\usepackage{colortbl}
\usepackage{pgfplots}
\pgfplotsset{compat=1.18}
\usepackage{siunitx} 
\usepackage{adjustbox}
\usepackage{placeins} 
\usepackage{float} 

\title{Real-time Cricket Sorting By Sex\\
\large\textit{A low-cost embedded solution using YOLOv8 and Raspberry Pi}}
\author[1,2]{Juan Manuel Cantarero Angulo}
\author[1]{Dr. Matthew Smith}
\affil[1]{Illinois Institute of Technology, Chicago, Illinois}
\affil[2]{Universidad Politécnica de Madrid, Madrid, Spain}
\date{}

\begin{document}

\twocolumn[
\begin{@twocolumnfalse}
\maketitle

\begin{abstract}
The global demand for sustainable protein sources is driving increasing interest in edible insects, with \textit{Acheta domesticus} (house cricket) identified as one of the most suitable species for industrial production. Current farming practices typically rear crickets in mixed-sex populations without automated sex sorting, despite potential benefits such as selective breeding, optimized reproduction ratios, and nutritional differentiation. This work presents a low-cost, real-time system for automated sex-based sorting of \textit{Acheta domesticus}, combining computer vision and physical actuation. The device integrates a Raspberry Pi 5 with the official Raspberry AI Camera and a custom YOLOv8 nano object detection model, together with a servo-actuated sorting arm. The model reached a mean Average Precision at IoU 0.5 ($mAP@0.5$) of 0.977 during testing, and real-world experiments with groups of crickets achieved an overall sorting accuracy of 86.8\%. These results demonstrate the feasibility of deploying lightweight deep learning models on resource-constrained devices for insect farming applications, offering a practical solution to improve efficiency and sustainability in cricket production.\\

\textit{\textbf{Keywords:} YOLO, Embedded computer vision, Insect farming, Object detection, Raspberry Pi, Real-time sorting, \textit{Acheta domesticus}, Edge AI}
\end{abstract}

\end{@twocolumnfalse}
]

\section{Introduction and Related Work}

The global demand for sustainable protein sources is rapidly increasing, driven by population growth, environmental concerns, and the limitations of conventional livestock production. Insects, in particular, have been highlighted by the FAO and numerous researchers as a highly efficient alternative, requiring significantly less land, water, and feed while producing lower greenhouse gas emissions \cite{fao}. Within this context, the insect protein market is projected to expand from approximately USD 2.4 billion in 2025 to about USD 28.54 billion by 2035, representing a compound annual growth rate of 28.1\% \cite{meticulous2025}.

Among edible insects, \textit{Acheta domesticus}, commonly known as the house cricket, has emerged as one of the leading candidates for industrial-scale farming due to its fast growth cycle, palatability, and acceptance across different cultures. In 2022, it was authorized as a novel food by the European Union (2022/188) \cite{eu}. Furthermore, \textit{Acheta domesticus} has been identified as one of the seven insect species suitable for industrial-scale production and human consumption \cite{vanHuis2017}.

Currently, house crickets are reared in mixed-sex populations under controlled environmental conditions \cite{farm}. However, no significant industrial efforts have been made to sort them by sex. Sex-based sorting could provide multiple advantages, including selective breeding, optimization of male-to-female ratios for reproduction, and nutritional improvements, as female crickets are generally larger \cite{size} and richer in protein \cite{protein} compared to males. Excess males, on the other hand, can generate competition, stress, and even cannibalistic behavior within colonies \cite{ratios}. Bridging this gap is critical to support the scalability of insect farming.

Previous research has been conducted on \textit{Acheta domesticus} gender identification and counting. For instance, Giulietti et al. \cite{giulietti2024vision} developed a vision-based measurement system to estimate the number and gender distribution of crickets using a test bench setup. Hansen et al. \cite{example_2} also applied state-of-the-art deep learning models (YOLOv5 for detection and VGG-16 for sex classification) to crickets, achieving high accuracy in sex identification but without integrating a physical sorting mechanism. However, both works focused solely on classification tasks, without providing a fully automated, real-time sorting pipeline.

In this paper, we go beyond these previous approaches by presenting a proof-of-concept prototype that not only classifies but also physically separates house crickets by sex in real time. The system integrates a YOLOv8 nano object detection model \cite{Jocher_Ultralytics_YOLO_2023} embedded on a Raspberry Pi 5 \cite{raspberrypi}, which controls a servo-actuated sorting arm. This design introduces the novel contribution of combining embedded computer vision with a mechanical sorting feature, enabling low-cost, autonomous operation suitable for small-scale insect farming and providing a foundation for more advanced industrial applications.

\section{Methodology}

The proposed system enables real-time classification and separation of AD by combining computer vision with embedded physical control. The hardware setup includes a Raspberry Pi 5, the official Raspberry Pi AI Camera, and a robotic arm actuated by a servo motor SG90. The software component is based on a lightweight object detection model (YOLOv8 nano), trained on a custom dataset, and integrated with control logic that activates the arm according to the predicted sex of the cricket.

The crickets are encouraged to move from an initial chamber (BOX 1) to a target chamber (BOX 2) via a transparent acrylic bridge. As a cricket crosses the bridge, the camera continuously captures its frames and sends them to the object detection model, which processes them in real time to determine both the cricket's position and sex. Based on these prediction, the system commands the servo motor to activate a flipper-like mechanism, directing the individual to either the male or female compartment within BOX 2.

\subsection{Hardware and Physical Setup}

\begin{figure}[!htbp]
  \centering
  \includegraphics[width=0.95\linewidth]{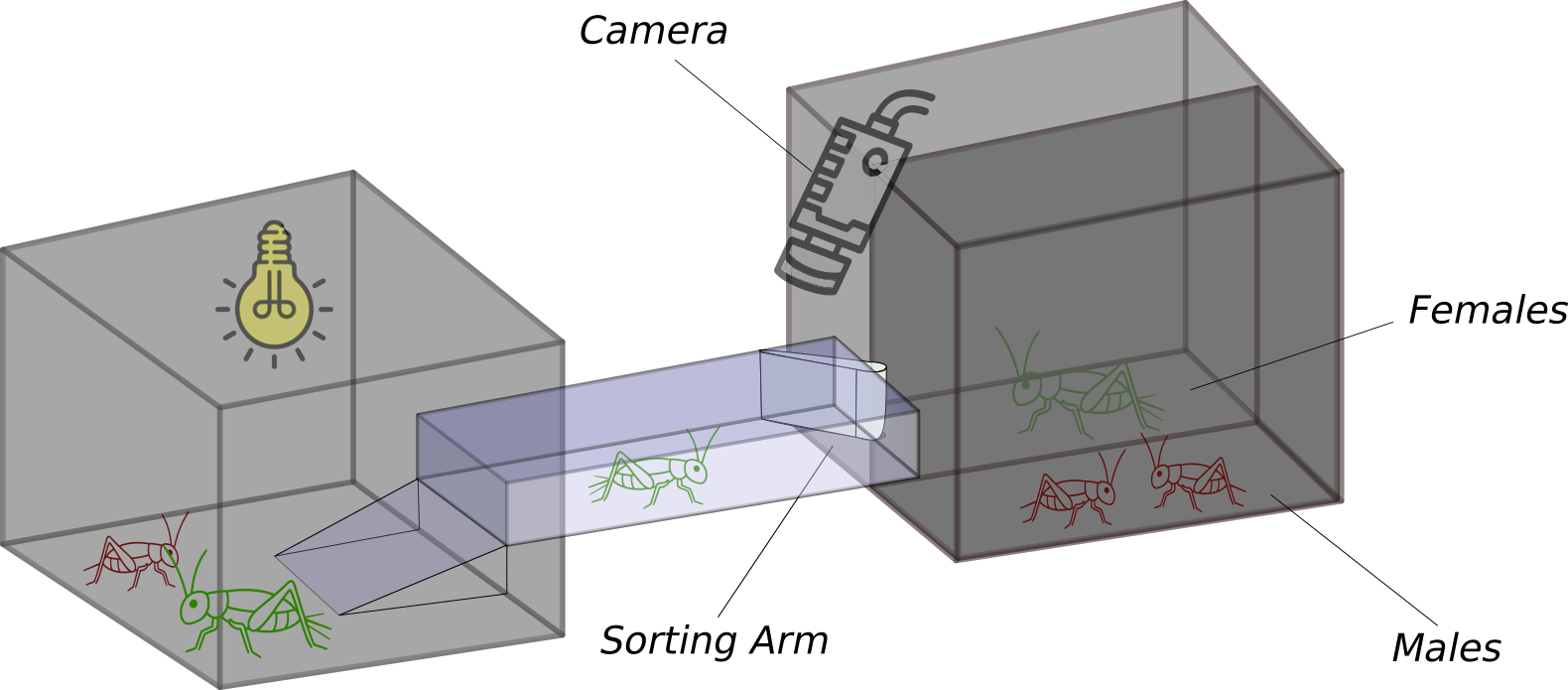}\\
  \includegraphics[width=0.95\linewidth]{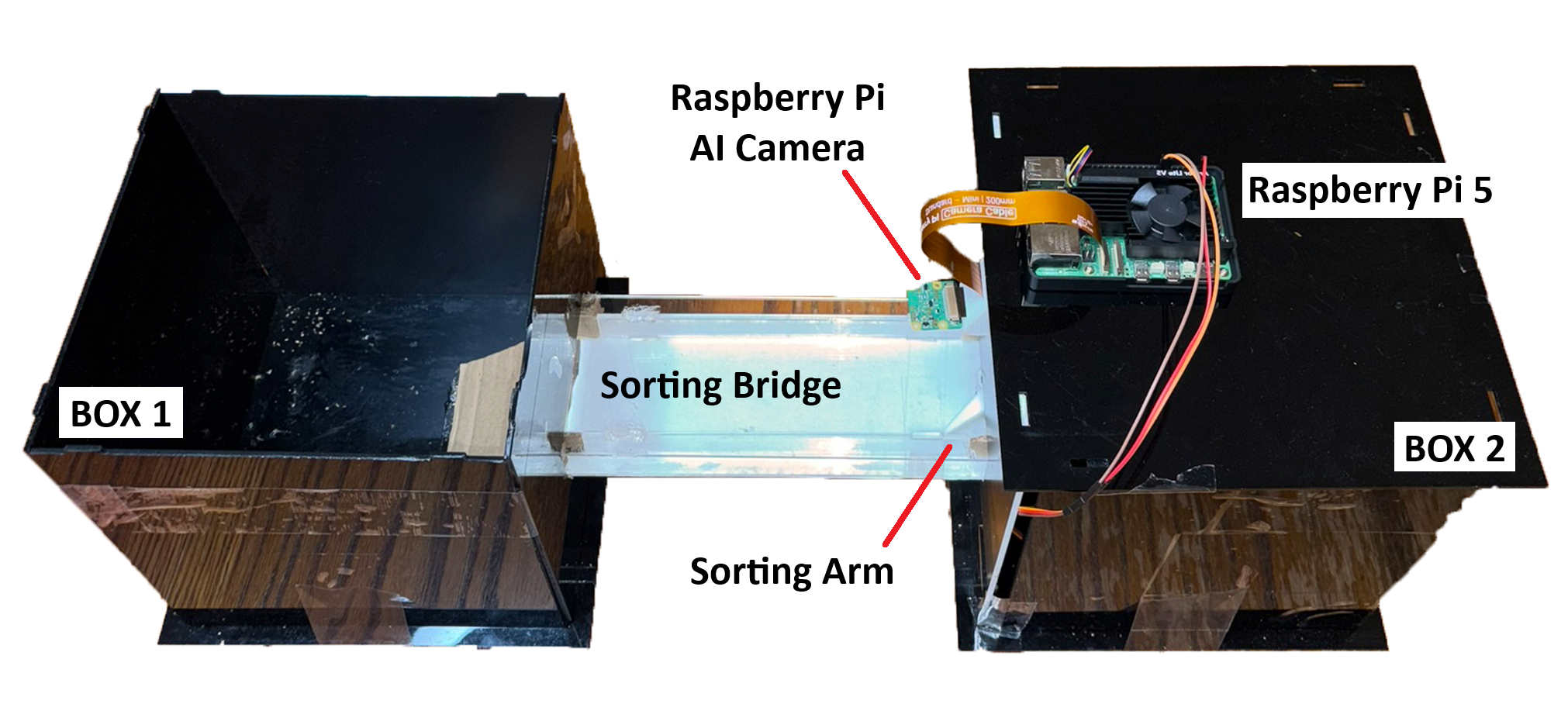}\\
  \includegraphics[width=0.8\linewidth]{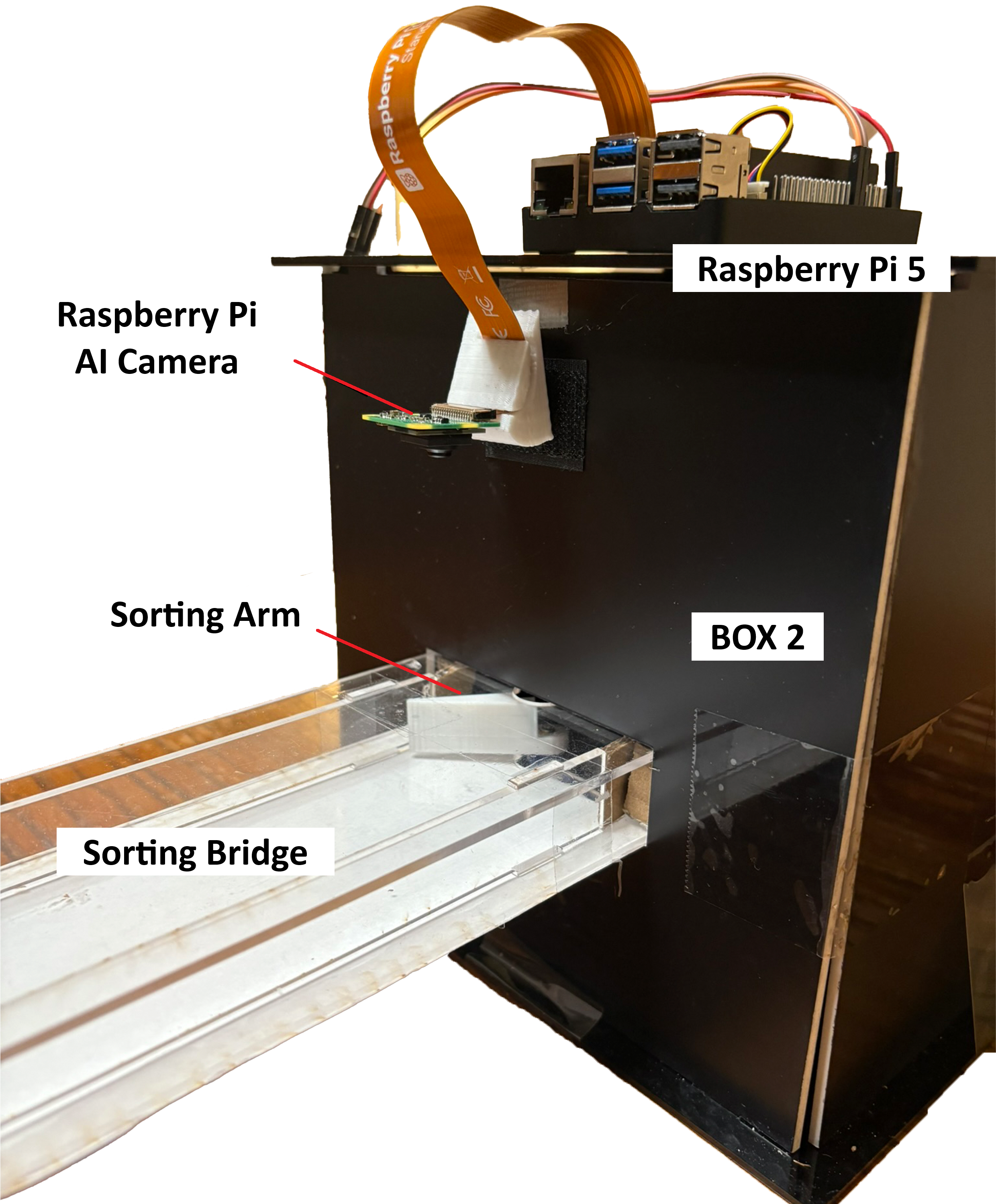}
  \caption{Overview of the cricket sorting prototype: \textbf{top} — drawing of the device; \textbf{center} — top view of the sorting device; \textbf{bottom} — Raspberry Pi AI camera and sorting arm.}
  \label{fig:prototype_overview}
\end{figure}

The walls of BOX 1 and BOX 2 are made of dark acrylic, while the connecting bridge is made of transparent acrylic (Figure \ref{fig:prototype_overview}); all components were designed in CAD and fabricated using a laser cutter. The sorting arm was also designed using CAD software and manufactured with a 3D printer.

A single camera (Raspberry Pi AI Camera) is mounted in a fixed overhead position above the transparent bridge, ash shown in Fig. \ref{fig:prototype_overview}, aligned with the sorting arm and the entrance to BOX 2. This placement allows the system to detect crickets just before they reach the sorting point, enabling timely activation of the arm. To enhance visual contrast and facilitate classification, the floor of the acrylic bridge has been painted white, providing a uniform background for image processing.

\begin{figure}[!htbp]
    \centering
    \includegraphics[width=0.8\linewidth]{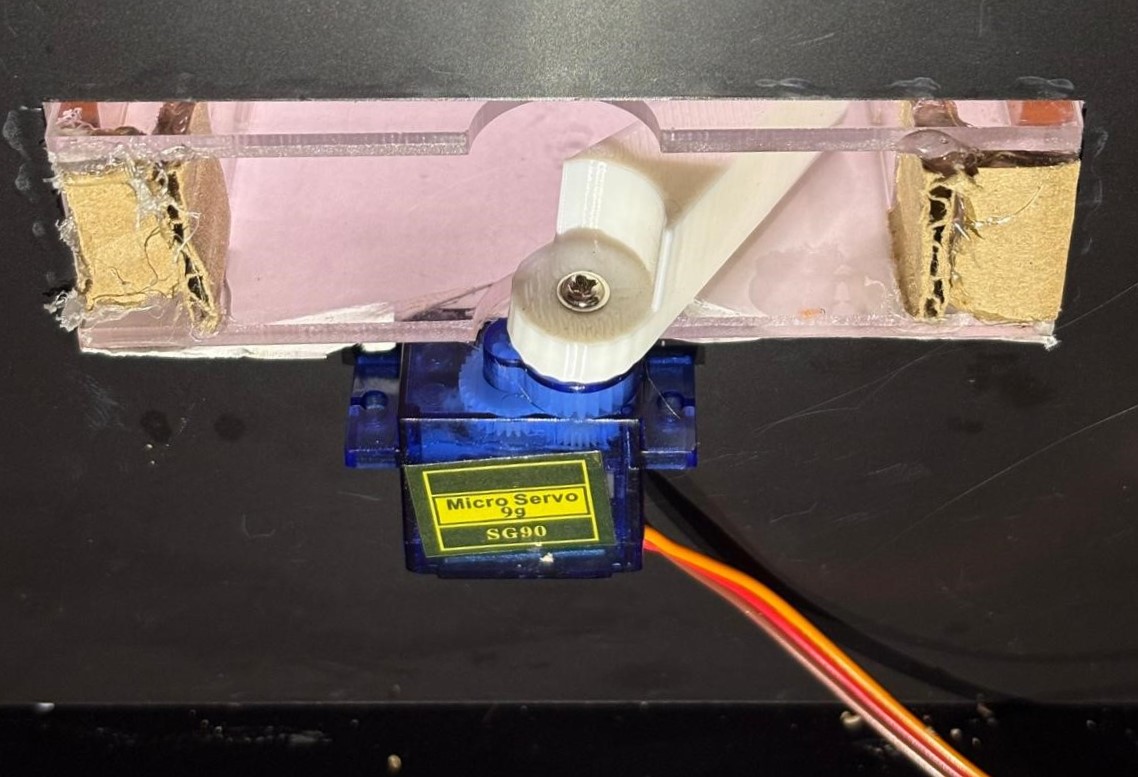}
    \caption{Servomotor and 3D-printed sorting arm located at the end of the bridge.}
    \label{servo}
\end{figure}

The sorting mechanism is driven by an SG90 servomotor (Fig. \ref{servo}), programmed to operate in two fixed positions (Fig. \ref{fig:sorting_arm_positions}). When the detection model identifies a female, the sorting arm blocks access to the male compartment, directing the cricket into the female section (FIg. \ref{fig:arm_female}). Conversely, if a male is detected, the arm switches to the opposite position (Fig. \ref{fig:arm_male}) . This non-invasive mechanism enables automated sorting without direct contact, minimizing the risk of injury to the insects.

To encourage the crickets to move through the system, two contrasting environments are created: BOX 1 contains an intense light source and lacks food and water, making it inhospitable, while BOX 2 remains dark and contains both food and moisture, thereby attracting the crickets naturally toward the destination chamber.

\subsection{Dataset and Annotation}

To train and validate the cricket detection model, a custom dataset was developed and made publicly available on Roboflow \cite{auto-crickets}. The dataset initially comprised 335 images with a resolution of 480$\times$480 pixels, capturing \textit{Acheta domesticus} individuals of both sexes as they approached the sorting arm. All images were acquired using the Raspberry Pi AI Camera, mounted in a fixed overhead position identical to that used during the experimental phase, ensuring consistency between data collection and deployment conditions.

\begin{figure}[!htbp]
    \centering
    \begin{subfigure}[t]{0.49\linewidth}
        \centering
        \includegraphics[width=\linewidth]{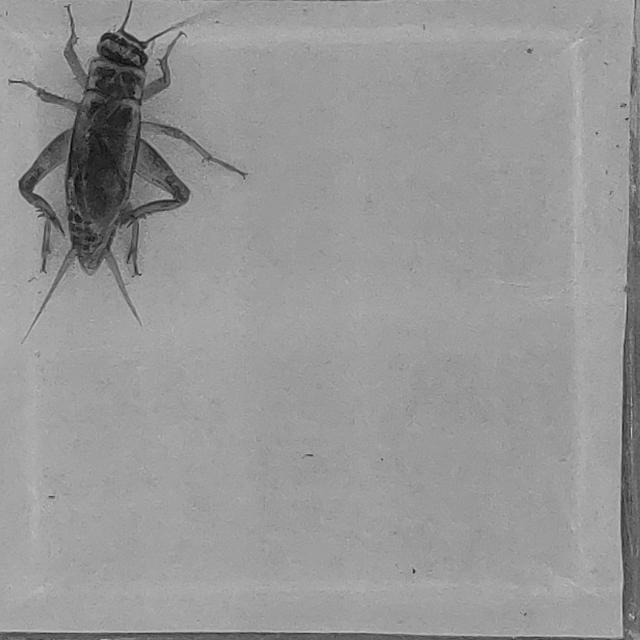}
        \subcaption{male}
        \label{fig:ad_male}
    \end{subfigure}\hfill
    \begin{subfigure}[t]{0.49\linewidth}
        \centering
        \includegraphics[width=\linewidth]{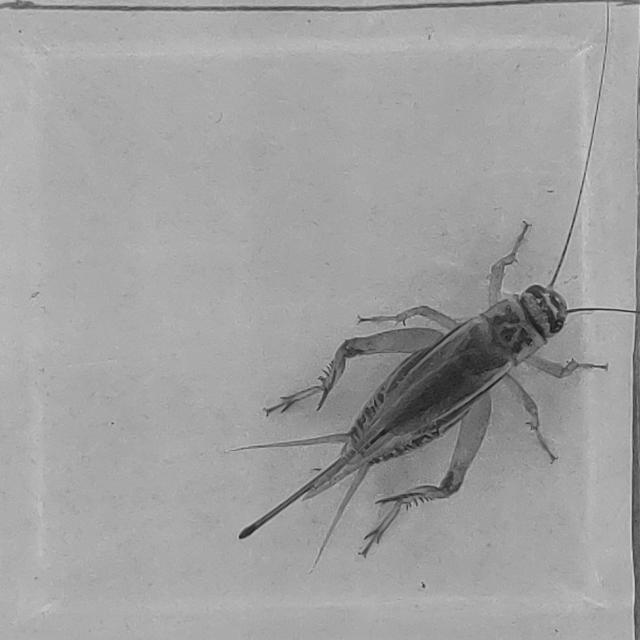}
        \subcaption{female}
        \label{fig:ad_female}
    \end{subfigure}
    \caption{Comparison between male and female \textit{Acheta domesticus}; note the ovipositor.}
    \label{fig:ad_mf_comparison}
\end{figure}

\begin{figure}[!htbp]
    \centering
    \includegraphics[width=0.8\linewidth]{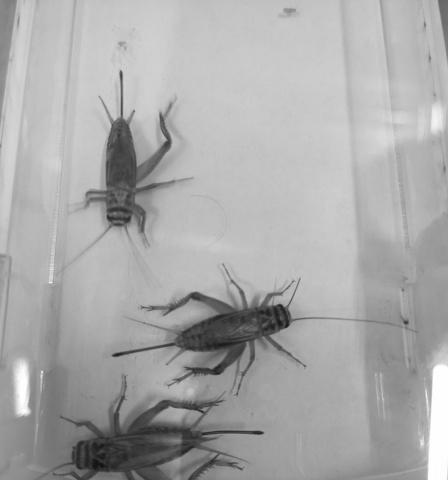}
    \caption{Training image containing three \textit{Acheta domesticus}.}
    \label{3_crickets}
\end{figure}

To ensure robustness under real-world conditions, various light sources were used during image acquisition, placed at different angles to compensate for reflections caused by the acrylic surface. This introduced controlled variability in illumination, helping the model generalize to different lighting scenarios. Additionally, images were taken with one, two, or even three (see figure \ref{3_crickets}) crickets of different sexes crossing the bridge simultaneously, increasing the diversity and complexity of the dataset.

\begin{figure}[h]
    \centering
    \includegraphics[width=1\linewidth]{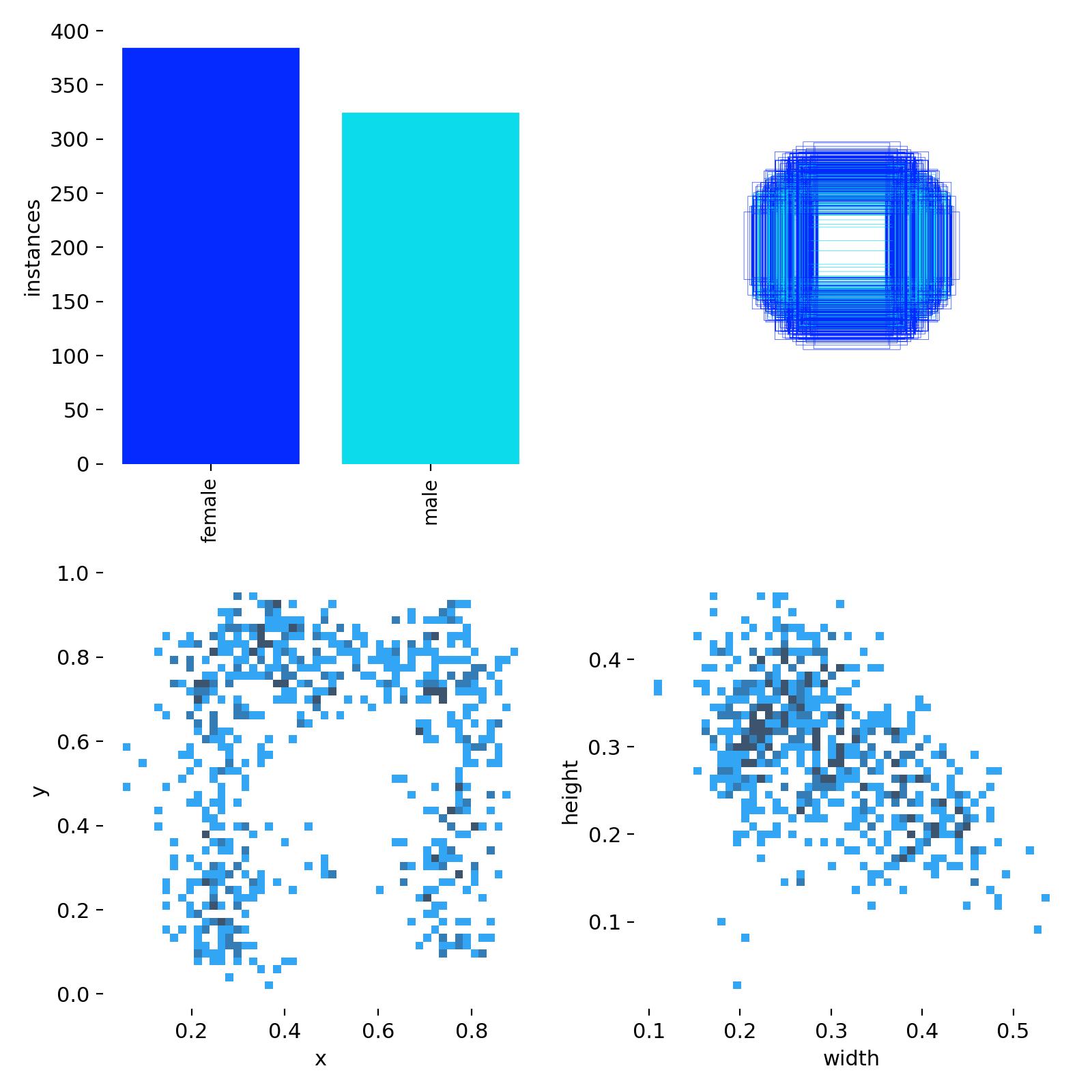}
    \caption{Spatial distribution of crickets in the dataset.}
    \label{position_size}
\end{figure}

Figure \ref{position_size} provides an overview of the dataset. It first highlights that the dataset is balanced, containing approximately the same number of male and female individuals. It also illustrates the spatial distribution of crickets within the images, showing that \textit{Acheta domesticus} often tend to walk close to the bridge walls, as well as their relative sizes (normalized between 0 and 1). This variability in both position and scale contributes to the robustness of the dataset and supports the expectation of reliable training outcomes.

Each image was converted to grayscale to reduce the input tensor to a single channel, thereby optimizing inference speed without significantly impacting detection accuracy. Following that, all images were annotated manually using Roboflow \cite{roboflow2025}. For each cricket, a bounding box was drawn and labeled according to its sex. Sex identification is straightforward to the human eye, with females being generally larger and exhibiting a clearly visible ovipositor at the rear end of the abdomen, as shown in figure \ref{fig:ad_mf_comparison}.

\begin{figure}[h]
    \centering
    \includegraphics[width=0.8\linewidth]{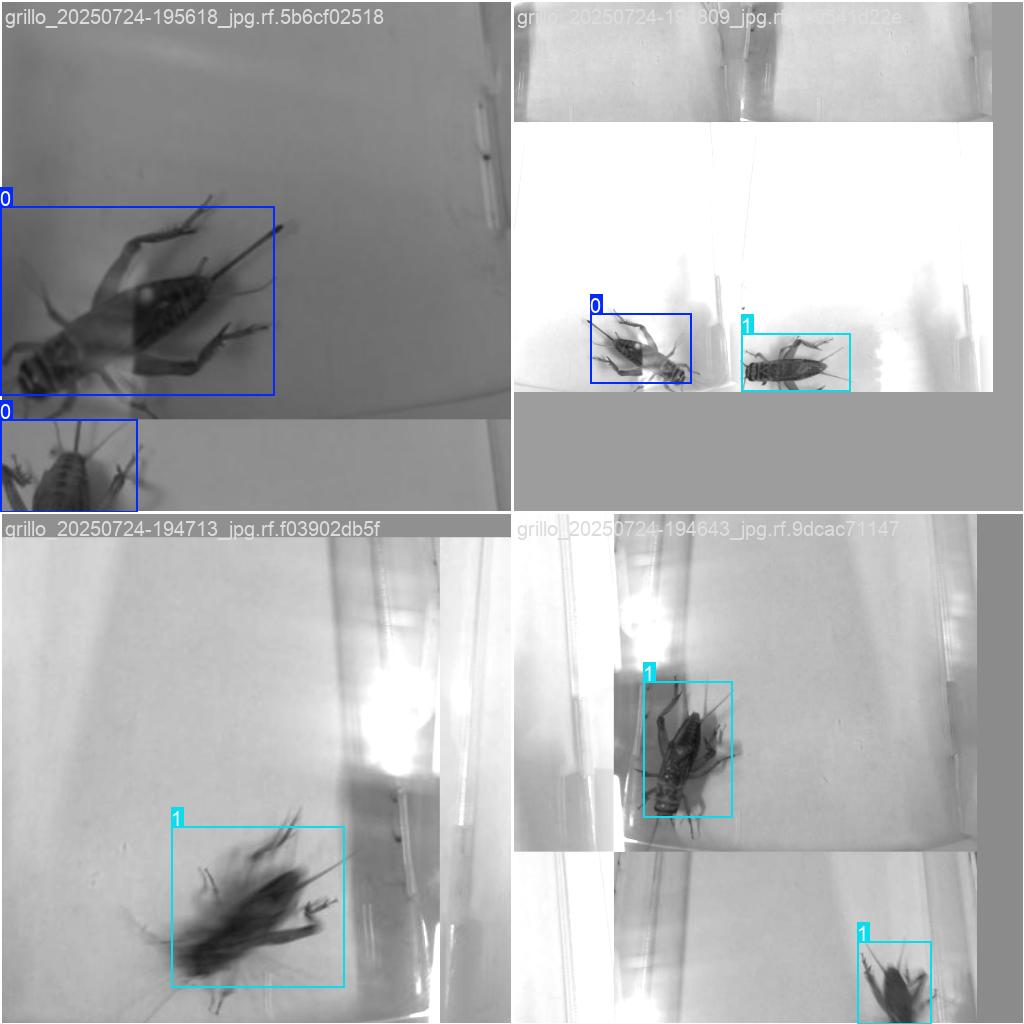}
    \caption{Training images.}
    \label{test1}
\end{figure}

To increase the size and variability of the dataset, a data augmentation process was applied. A random zoom was introduced in the range of 10\% to 26\% to simulate partial occlusions, such as cases where a cricket is only partially visible due to the frame or overlapping individuals. After augmentation, the final dataset contained 597 training images (81\%), 63 validation images (9\%), and 73 test images (10\%), all exported in YOLOv8 format with normalized bounding box coordinates.

\FloatBarrier
\subsection{Object Detection Model: YOLOv8 Nano}
The object detection model employed in this work is YOLOv8 nano, the lightest variant of the YOLOv8 family developed by Ultralytics \cite{Jocher_Ultralytics_YOLO_2023}. YOLO models are single-ahot detectors (SSDs) capable of predicting bounding boxes and their associated classes in a single forward pass through the neural network. This design eliminates the need for intermediate region proposal steps, resulting in significantly reduced inference times compared to two-stage detectors such as Faster R-CNN.

YOLOv8 introduces several enhancements over previous versions, including decoupled heads for regression and classification, and a redesigned backbone based on CSPDarknet \cite{bochkovskiy2020yolov4} with Cross Stage Partial (CSP) connections. The nano version is specifically optimized for deployment on resource-constrained devices by reducing the depth and width multipliers of the network, thereby decreasing both the number of parameters and the computational load. This enables real-time execution on the Raspberry Pi 5 CPU without the need for external GPU acceleration.

In this work, YOLOv8 nano was trained locally on the grayscale dataset described before. During inference, the model processes images captured by the Raspberry Pi AI Camera and, for each frame, outputs the coordinates of the detected bounding box(es), the predicted class (male or female) for each box, and the corresponding confidence score. If there is no cricket detected on the frame, then the model returns: "no detection".

Object detection model performance is commonly evaluated using 
\textit{Intersection over Union (IoU)} and 
\textit{mean Average Precision (mAP)}. IoU quantifies the overlap between the predicted box 
$B_{p}$ and the ground truth box $B_{gt}$:

\begin{equation}
IoU = \frac{|B_{p} \cap B_{gt}|}{|B_{p} \cup B_{gt}|}
\end{equation}

A detection is considered correct when $IoU \geq 0.5$. The \textit{Average Precision (AP)} of a class is the area under 
its Precision--Recall curve:

\begin{equation}
AP_{c} = \int_{0}^{1} P_{c}(R)\, dR
\end{equation}

and the \textit{mAP} is the mean of all AP classes. In this work, 
results are reported as $mAP@0.5$, i.e., detections are correct if 
$IoU \geq 0.5$.

\begin{figure}[!htbp]
    \centering
    \includegraphics[width=0.8\linewidth]{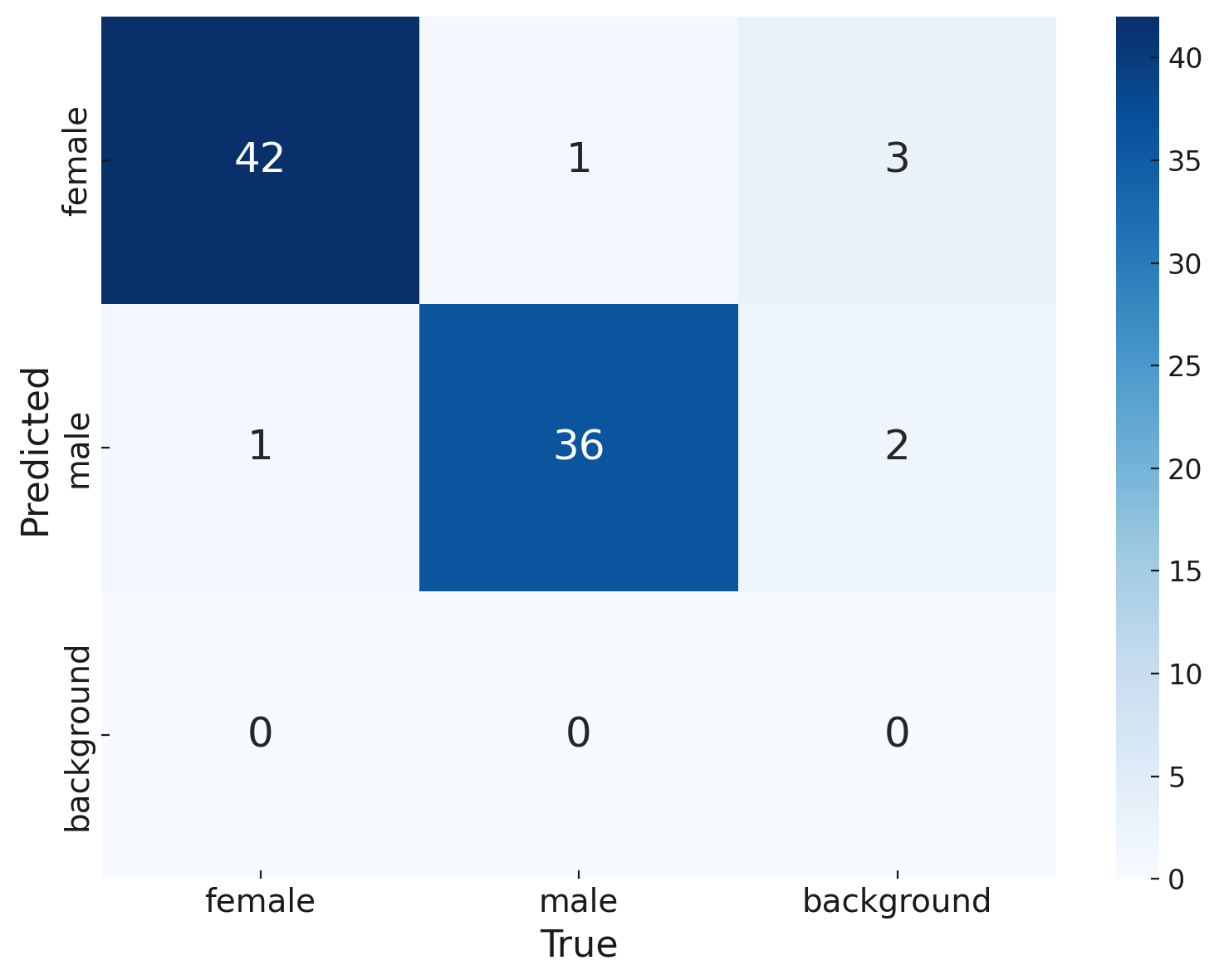}
    \caption{Confusion matrix obtained from test images.}
    \label{conf_test}
\end{figure}

\begin{figure}[!htbp]
    \centering
    \includegraphics[width=1\linewidth]{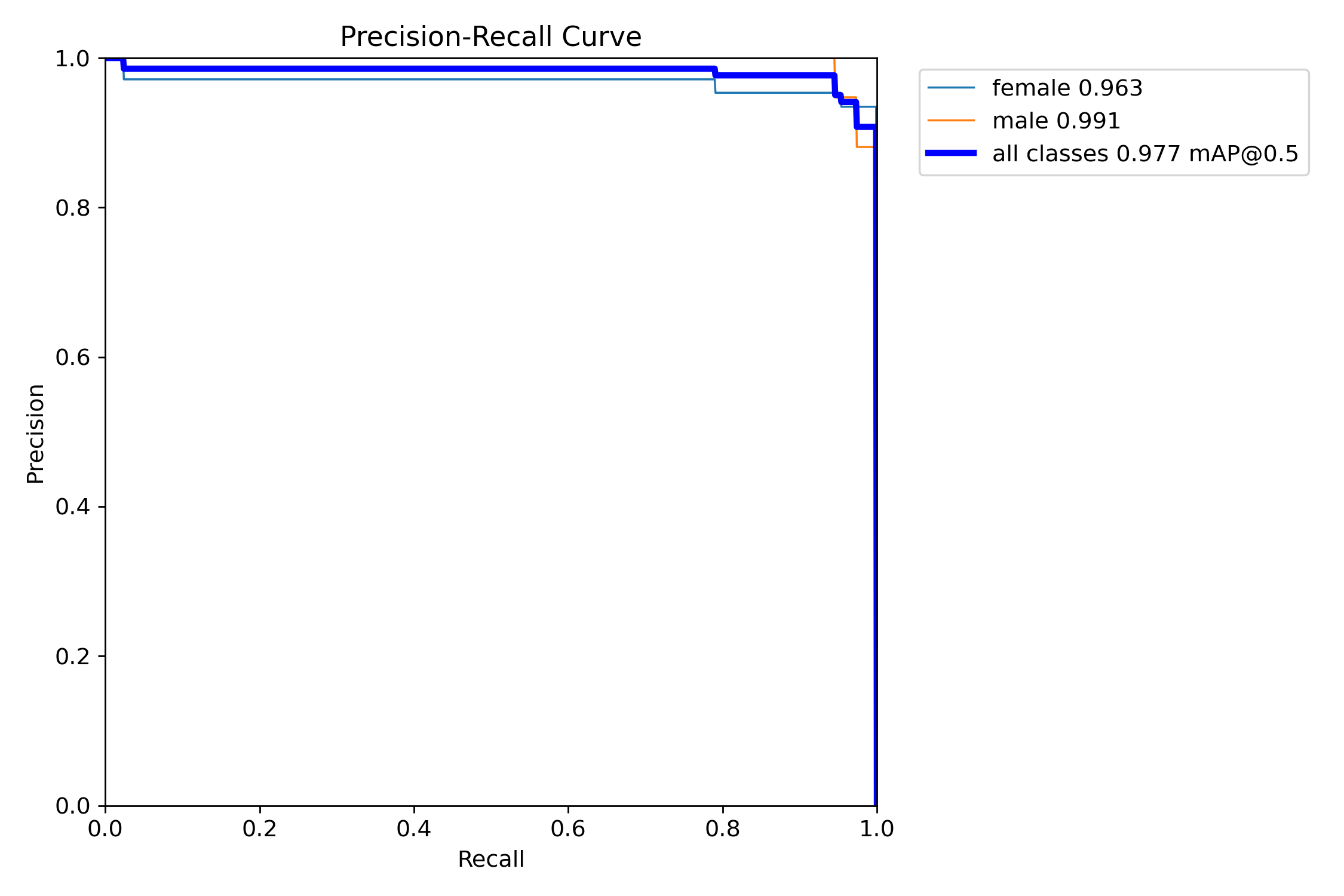}
    \caption{Precision--Recall curve of the trained YOLO model during test.}
    \label{pr_curve}
\end{figure}

Training results are highly promising. Figure \ref{conf_test} presents the confusion matrix obtained during the test phase, while Figure \ref{pr_curve} shows the Precision–Recall curve used to calculate the model's $mAP@0.5$, which reached 0.977. Misclassification was minimal: only 1 out of 43 females (~2.3\%) was classified as male, and 1 out of 37 males (~2.7\%) as female, resulting in an overall accuracy of 97.5\%. Nevertheless, the model produced five false positives in background regions with no crickets present. These errors are likely due to subtle variations in lighting conditions and could be mitigated by expanding the dataset with a larger and more diverse set of images.

\FloatBarrier
\section{Control Logic and Actuation}

The system is not limited to determining the sex of each cricket; it also incorporates control logic to decide when to activate the sorting mechanism, thereby reducing the likelihood of incorrect sorting actions.

\begin{figure}[!htbp]
    \centering
    \includegraphics[width=0.8\linewidth]{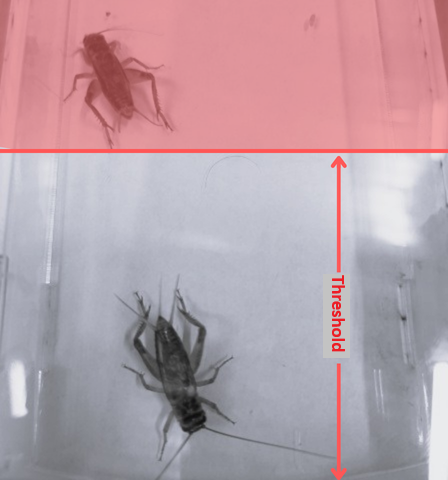}
    \caption{Distance threshold. The upper cricket is ignored.}
    \label{threshold}
\end{figure}


The sorting arm is triggered only when the cricket is sufficiently close to the mechanism. As multiple crickets may appear in the same frame, the system prioritizes the individual closest to the sorting point. For this purpose, a fixed proximity threshold is defined. If the detected cricket is located beyond this threshold, the arm remains inactive. This threshold is visually represented by a horizontal reference line in Figure~\ref{threshold}, which indicates the minimum detection distance required to activate the actuator (100 pixels in this case).

\begin{figure}[h]
  \renewcommand{\thefigure}{S1} 
  \centering
  \begin{subfigure}[t]{0.48\linewidth}
    \centering
    \includegraphics[width=\linewidth]{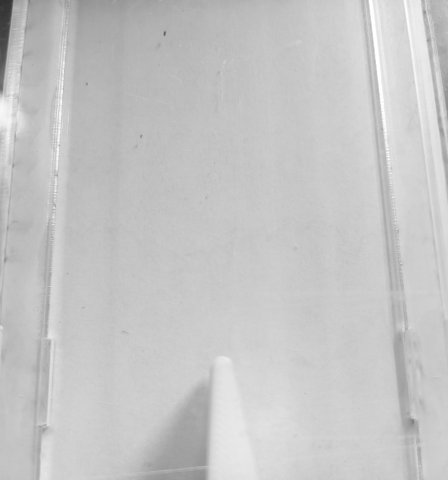}
    \subcaption{Neutral position}
    \label{fig:arm_neutral}
  \end{subfigure}\hfill
  \begin{subfigure}[t]{0.48\linewidth}
    \centering
    \includegraphics[width=\linewidth]{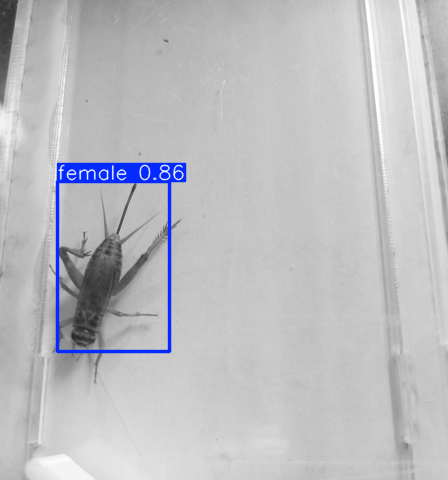}
    \subcaption{Female sorting position}
    \label{fig:arm_female}
  \end{subfigure}

  \vspace{0.3cm} 

  \begin{subfigure}[t]{0.48\linewidth}
    \centering
    \includegraphics[width=\linewidth]{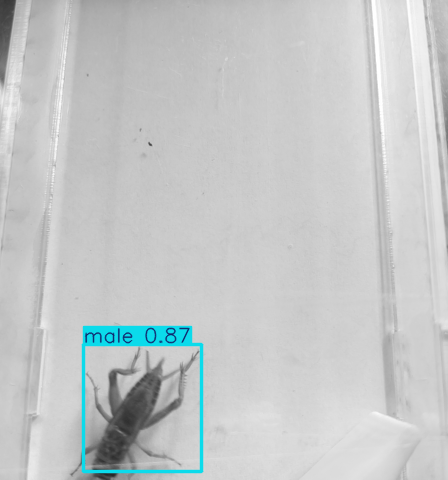}
    \subcaption{Male sorting position}
    \label{fig:arm_male}
  \end{subfigure}

  \caption{Sorting arm positions.}
  \label{fig:sorting_arm_positions}
\end{figure}
\renewcommand{\thefigure}{\arabic{figure}} 

To prevent unintended activations caused by occasional misclassifications —an inherent challenge in object detection systems— a sliding window of 10 frames is also applied. With an acquisition rate of approximately one frame every 300 ms, the window spans roughly three seconds. The system computes an average prediction over these 10 frames; the arm is activated only if the majority of them meet both the class and proximity criteria.

Figure \ref{fig:sorting_arm_positions} illustrates the three possible positions of the sorting arm: neutral (initial), female sorting position, and male sorting position.

\FloatBarrier
\section{Experiments and Results}

Four experiments were conducted, each involving approximately 25 \textit{Acheta domesticus}. The objective of these real-time sorting trials was to assess the overall efficiency of the device under different stress conditions. In all cases, BOX 1 was illuminated with an intense light source and contained no food or water, while BOX 2 was kept dark and provided with food and water, thus attracting the crickets.

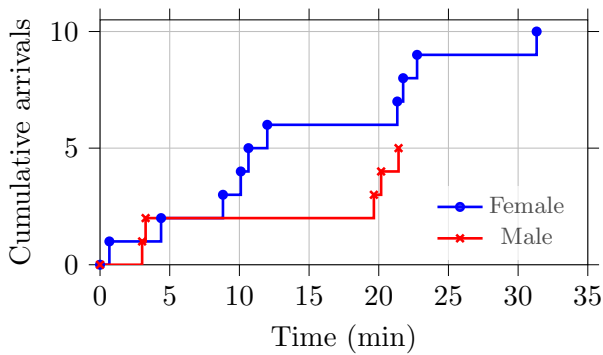
\begin{figure}[htbp]
  \centering
  \begin{tikzpicture}
    \begin{axis}[
      width=\columnwidth,
      height=0.6\columnwidth,
      xlabel={Time (min)},
      ylabel={Cumulative arrivals},
      xmin=0, xmax=35,
      ymin=0, ymax=10.5,
      grid=both,
      legend style={at={(0.98,0.02)},anchor=south east,draw=none,fill=white,fill opacity=0.7,font=\footnotesize},
      tick align=outside, tick style={black},
    ]
      \addplot+[mark=*, line width=1pt, mark options={scale=0.7}, const plot mark right, const plot]
      coordinates {
        (0,0)
        (0.67,1) (4.38,2) (8.82,3)
        (10.10,4) (10.65,5) (12.00,6)
        (21.33,7) (21.75,8) (22.75,9) (31.33,10)
      };
      \addlegendentry{Female}

      \addplot+[mark=x, line width=1pt, mark options={scale=1}, const plot mark right, const plot]
      coordinates {
        (0,0)
        (3.02,1) (3.27,2)
        (19.65,3) (20.17,4) (21.42,5)
      };
      \addlegendentry{Male}
    \end{axis}
  \end{tikzpicture}
  \caption{Experiment~3: cumulative arrivals by sex.}
  \label{fig:cum_exp3}
\end{figure}

In Experiment~1, crickets were manually forced to cross the bridge one by one. This procedure generated high stress, causing them to move nervously and at high speed. As a result, in some cases the YOLOv8 model detected the cricket correctly, but the sorting arm could not react fast enough, leading to misclassifications. In this context, we define the \textit{sorting accuracy} as the accuracy of the entire system, which includes both the vision-based classification and the mechanical actuation of the sorting arm. It should not be confused with the object detection accuracy of the classifier alone. Sorting accuracy is calculated as the ratio between the number of crickets that were successfully separated into the correct compartment by the arm and the total number of crickets sorted, i.e.,
\[
\text{Sorting Acc.} = \frac{\text{Number of correctly sorted crickets}}{\text{Total number of sorted crickets}}.
\]

\begin{figure}[htbp]
  \centering
  \begin{tikzpicture}
    \begin{axis}[
      width=\columnwidth,
      height=0.6\columnwidth,
      xlabel={Time (min)},
      ylabel={Cumulative arrivals},
      xmin=0, xmax=110,
      ymin=0, ymax=15.5,
      grid=both,
      legend style={at={(0.98,0.02)},anchor=south east,draw=none,fill=white,fill opacity=0.7,font=\footnotesize},
      tick align=outside, tick style={black},
    ]
      \addplot+[mark=*, line width=1pt, mark options={scale=0.7}, const plot mark right, const plot]
      coordinates {
        
        (3.75,1) (5.67,2) (6.33,3) (8.58,4) (8.67,5)
        (11.33,6) (19.00,7) (19.42,8) (20.83,9) (22.50,10)
        (40.83,11) (50.00,12) (65.07,13) (84.83,14) (109.75,15)
      };
      \addlegendentry{Female}

      \addplot+[mark=x, line width=1pt, mark options={scale=1}, const plot mark right, const plot]
      coordinates {
        
        (19.50,1) (22.30,2) (52.42,3)
        (71.50,4) (71.80,5) (72.17,6)
        (107.28,7) (108.97,8) (108.98,9) (109.28,10)
      };
      \addlegendentry{Male}
    \end{axis}
  \end{tikzpicture}
  \caption{Experiment~4: cumulative arrivals by sex.}
  \label{fig:cum_exp4}
\end{figure}

\FloatBarrier

For Experiment~1, the sorting accuracy achieved was 83\%, considerably lower than the 94\% obtained in Experiment~3. Although performance was lower compared to other scenarios, the main advantage of this approach was speed: all crickets were sorted in only 10 minutes. However, its major drawback was the requirement of constant human intervention, which makes the method unsuitable for an autonomous and scalable system.

\begin{table}[!htbp]
\centering
\caption{Confusion matrix for Experiment~1.}
\label{tab:conf_exp1}
\begin{tabular}{c|cc}
\hline
 & Pred. Male & Pred. Female \\
\hline
True Male   & 10 & 2 \\
True Female & 2  & 9 \\
\hline
\end{tabular}
\end{table}

In Experiment~2, instead of pushing the crickets individually, BOX 1 was shaken to induce collective movement. This induced in \textit{Acheta domesticus} a nervousness very similar to that obtained in Experiment~1. The classification results (Table \ref{tab:conf_exp2}) were similar to those in Experiment~1, but with a key advantage: box agitation can be automated, thereby avoiding the need for human interaction. The total duration of the experiment was 15 minutes, very similar to the 10 minutes of Experiment~1. As shown in table \ref{tab:metrics_total}, the accuracy obtained was 83\%

\begin{table}[!htbp]
\centering
\caption{Confusion matrix for Experiment~2.}
\label{tab:conf_exp2}
\begin{tabular}{c|cc}
\hline
 & Pred. Male & Pred. Female \\
\hline
True Male   & 13 & 1 \\
True Female & 4 & 11 \\
\hline
\end{tabular}
\end{table}

In Experiment~3, only occasional stimuli were applied, consisting of light agitation of BOX 1 and gentle air blows. This reduced the stress level of the crickets, allowing them to cross the bridge more calmly. Consequently, sorting was smoother, as reflected in Table \ref{tab:conf_exp3}. However, the reduced pace significantly increased the duration, requiring 35 minutes for all individuals to cross.

\begin{table}[!htbp]
\centering
\caption{Confusion matrix for Experiment~3.}
\label{tab:conf_exp3}
\begin{tabular}{c|cc}
\hline
 & Pred. Male & Pred. Female \\
\hline
True Male   & 5 & 0 \\
True Female & 1 & 11 \\
\hline
\end{tabular}
\end{table}

In Experiment~4, no agitation was applied, and the crickets were allowed to cross at their own pace. As shown in Table \ref{tab:conf_exp4}, this low-stress configuration produced the best classification metrics, but at the expense of time efficiency. After 1h 50 minutes, only 72\% of the crickets had crossed, and the experiment was concluded. Under these conditions not all individuals could be guaranteed to cross the bridge. The cumulative arrivals (Fig. \ref{fig:cum_exp4}) reveal a progressively declining crossing rate: most movements occurred during the first 30 minutes, while the remaining individuals delayed or entirely avoided crossing. This suggests that the more crickets remain in the starting box, the lower their tendency to move, possibly due to comfort, group effects, or reluctance to explore.

\begin{table}[!htbp]
\centering
\caption{Confusion matrix for Experiment~4.}
\label{tab:conf_exp4}
\begin{tabular}{c|cc}
\hline
 & Pred. Male & Pred. Female \\
\hline
True Male   & 8 & 1 \\
True Female & 1 & 13 \\
\hline
\end{tabular}
\end{table}

\begin{table}[!htbp]
\centering
\caption{Metrics summary.}
\label{tab:metrics}
\begin{tabular}{lccccc}
\hline
\textbf{Exp} & \textbf{Acc} & \textbf{Prec} & \textbf{Rec} & \textbf{F1} & \textbf{Time (min)} \\
\hline
1 & 0.83 & 0.83 & 0.83 & 0.83 & 10 \\
2 & 0.83 & 0.84 & 0.83 & 0.83 & 15 \\
3 & 0.94 & 0.92 & 0.96 & 0.93 & 35 \\
4 & 0.91 & 0.91 & 0.91 & 0.91 & 110 \\
\hline
\end{tabular}
\end{table}

\begin{figure}[h]
  \centering
  \includegraphics[width=1.0\columnwidth]{images/speed_boxplot.png}
  \caption{Distribution of cricket speeds (mm/s) by sex and classification outcome. Misclassified individuals of both sexes exhibit higher median speeds compared to correctly classified ones.}
  \label{fig:speed_boxplot}
\end{figure}

\begin{table}[!htbp]
\centering
\caption{Measured speed of crickets in Experiments 3 and 4.}
\label{tab:cricket_speeds}
\begin{adjustbox}{max width=\linewidth}
\begin{tabular}{lcccc}
\hline
\textbf{Cricket} & \textbf{Sex} & \textbf{Time (s)} & \textbf{Classified} & \textbf{Speed (mm/s)} \\
\hline
1  & f & 7.00  & f & 10.71 \\
2  & f & 4.00  & f & 18.75 \\
3  & f & 1.80  & f & 41.67 \\
4  & f & 2.05  & f & 36.59 \\
5  & f & 1.90  & f & 39.47 \\
6  & f & 1.10  & f & 68.18 \\
7  & m & 0.80  & f & 93.75 \\
8  & f & 4.10  & f & 18.29 \\
9  & f & 6.00  & f & 12.50 \\
10 & m & 5.10  & m & 14.71 \\
11 & f & 4.00  & f & 18.75 \\
12 & f & 4.10  & f & 18.29 \\
13 & f & 1.80  & f & 41.67 \\
14 & m & 1.80  & m & 41.67 \\
15 & f & 5.10  & f & 14.71 \\
16 & m & 1.90  & m & 39.47 \\
17 & m & 2.00  & m & 37.50 \\
18 & m & 1.95  & m & 38.46 \\
19 & f & 1.80  & f & 41.67 \\
20 & m & 2.10  & m & 35.71 \\
21 & m & 0.20  & m & 375.00 \\
22 & m & 0.70  & m & 107.14 \\
23 & m & 1.00  & m & 75.00 \\
24 & f & 0.10  & m & 750.00 \\
25 & f & 1.00  & f & 75.00 \\
26 & m & 4.50  & m & 16.67 \\
27 & m & 3.10  & m & 24.19 \\
28 & f & 0.80  & f & 93.75 \\
29 & f & 11.00 & f & 6.82 \\
30 & f & 2.10  & f & 35.71 \\
31 & f & 7.00  & f & 10.71 \\
32 & f & 10.00 & f & 7.50 \\
33 & f & 8.50  & f & 8.82 \\
34 & m & 6.60  & m & 11.36 \\
35 & m & 3.80  & m & 19.74 \\
36 & f & 4.00  & m & 18.75 \\
37 & m & 3.70  & m & 20.27 \\
38 & f & 20.00 & f & 3.75 \\
39 & f & 5.60  & f & 13.39 \\
40 & f & 8.10  & f & 9.26 \\
\hline
\end{tabular}
\end{adjustbox}
\end{table}

\FloatBarrier

To further investigate the conditions that affected sorting performance, we analyzed the speed of individual crickets in relation to their true sex and whether they were correctly or incorrectly classified by the system (Fig.~\ref{fig:speed_boxplot}). The speeds were obtained by measuring the time elapsed from the moment each cricket appeared in the detection zone of the camera’s field of view until it completely crossed the bridge, and dividing that time by the length of the traversed section (Table \ref{tab:cricket_speeds}). The results reveal two main trends. First, males were on average faster than females (median male speed: 39.5~mm/s; median female speed: 18.8~mm/s). Second, misclassified individuals, regardless of sex, tended to move at higher speeds than those correctly sorted (median misclassified speed: 75.0~mm/s vs. 20.3~mm/s for correctly classified). These findings suggest that higher movement speed challenged the mechanical response of the sorting arm, leading to incorrect classifications. 

\begin{table}[!htbp]
\centering
\caption{Global confusion matrix obtained by summing the four real-time sorting experiments.}
\label{tab:conf_total}
\begin{tabular}{c|cc}
\hline
 & Pred. Male & Pred. Female \\
\hline
True Male   & 36 & 6 \\
True Female & 6  & 43 \\
\hline
\end{tabular}
\end{table}

\begin{table}[!htbp]
\centering
\caption{Class-wise performance metrics derived from the global confusion matrix.}
\label{tab:metrics_total}
\begin{tabular}{l|cc}
\hline
 & Male & Female \\
\hline
Precision   & 0.857 & 0.878 \\
Recall      & 0.857 & 0.878 \\
F1-score    & 0.857 & 0.878 \\
Specificity & 0.878 & 0.857 \\
Accuracy    & 0.868 & 0.868 \\
\hline
\end{tabular}
\end{table}

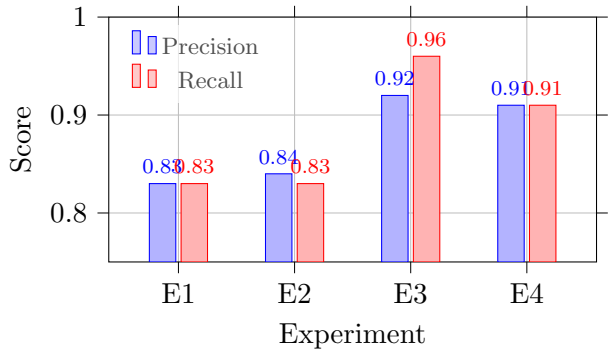
\begin{figure}[!htbp]
  \centering
  \begin{tikzpicture}
    \begin{axis}[
      width=\columnwidth,
      height=0.6\columnwidth,
      ybar,
      bar width=10pt,
      ymin=0.75, ymax=1.0,
      symbolic x coords={E1,E2,E3,E4},
      xtick=data,
      xlabel={Experiment},
      ylabel={Score},
      grid=both,
      tick align=outside,
      tick style={black},
      legend style={
        at={(0.03,0.97)}, 
        anchor=north west,
        draw=none,
        fill=white,
        fill opacity=0.7,
        font=\footnotesize
      },
      enlarge x limits=0.2,
      nodes near coords,
      nodes near coords align={vertical},
      every node near coord/.append style={font=\scriptsize, /pgf/number format/precision=2},
    ]
      \addplot coordinates {(E1,0.83) (E2,0.84) (E3,0.92) (E4,0.91)};
      \addlegendentry{Precision}

      \addplot coordinates {(E1,0.83) (E2,0.83) (E3,0.96) (E4,0.91)};
      \addlegendentry{Recall}
    \end{axis}
  \end{tikzpicture}
  \caption{Precision and recall per experiment.}
  \label{fig:prec_rec_by_exp}
\end{figure}

\FloatBarrier

Table \ref{tab:conf_total} presents the global confusion matrix obtained by aggregating the four real-time experiments. This consolidated view provides a comprehensive picture of the device's overall performance across different conditions, resulting in an overall sorting accuracy of 86.8\%

\section{Conclusions and Future Work}

This work demonstrates the feasibility of a low-cost and autonomous device for real-time sex sorting of \textit{Acheta domesticus}, integrating computer vision and embedded control. The experiments revealed that, under low-stress conditions, the system achieved accuracies above 90\%, while high-stress conditions allowed faster throughput at the cost of lower classification performance.
 
To the best of our knowledge, this is the first prototype capable of physically separating house crickets by sex in real time. Such a system has the potential to contribute significantly to the rapidly expanding insect farming industry by enabling selective breeding, optimizing sex ratios for reproduction, and adjusting nutritional strategies.

Limitations were identified, including the influence of environmental stressors, light reflections from the acrylic setup, and the noise of the servo motor, which occasionally discouraged crickets from crossing. A further observation from Experiments 3 and 4 is that, due to the calm state of the crickets, some were startled by the noise of the sorting arm while crossing the bridge, often retreating back into BOX 1 and delaying the process. Interestingly, males appeared to be more skittish than females: in the last two experiments, where crickets were allowed to circulate freely, males were consistently the last to cross the bridge, seemingly more affected by the sound of the sorting arm.

Future work will focus on improving hardware robustness and scalability and testing alternative actuators. Moreover, the disturbing method applied to stimulate movement can be further automated, for example by introducing controllable heat pads in the base of BOX 1 or small fans, avoiding human interventions. Additional experiments with larger populations of crickets will also be necessary to increase the reliability and statistical significance of the results. Finally, extending the approach to other insect species and evaluating energy efficiency metrics will further explore the industrial potential of automated insect sorting.

\bibliography{bib}
\clearpage
\appendix
\section*{APPENDIX}

The dataset and source code supporting the findings of this work are publicly available on Roboflow and GitHub \cite{auto-crickets, cantarero2025github}.

\subsection*{NOMENCLATURE}
\begin{itemize}
\item[AD] \textit{Acheta domesticus}
\item[YOLO] You Only Look Once
\item[IoU] Intersection over Union
\item[mAP] mean Average Precision
\item[R-CNN] Region-based Convolutional Neural Network
\item[$B_{p}$] Predicted bounding box
\item[$B_{gt}$] Ground truth bounding box
\item[AP] Average Precision
\item[VGG] Visual Geometry Group
\item[SSD] Single-Shot Detector
\end{itemize}

\end{document}